\documentclass[10pt,twocolumn,letterpaper]{article}

\usepackage{iccv}
\usepackage{times}
\usepackage{epsfig}
\usepackage{graphicx}
\usepackage{amsmath}
\usepackage{amssymb}
\usepackage{booktabs}
\usepackage{multicol}
\usepackage{algorithm,algorithmicx,algpseudocode}
\usepackage[accsupp]{axessibility}

\usepackage[pagebackref=true,breaklinks=true,letterpaper=true,colorlinks,bookmarks=false]{hyperref}

\iccvfinalcopy

\ificcvfinal\fi

\begin{document}

\title{Do DALL-E and Flamingo Understand Each Other?}

\author{{Hang Li$^*$}$^{1,2}$
\qquad
{Jindong Gu$^*$}$^{3}$
\qquad
{Rajat Koner}$^{1}$
\qquad
{Sahand Sharifzadeh}$^{1}$
\qquad
{Volker Tresp}$^{1,2}$\\
{$^{1}$LMU Munich, Germany\qquad $^{2}$Siemens AG, Germany\qquad $^{3}$University of Oxford, UK}\\
}

\maketitle
\def\thefootnote{*}\footnotetext{Equal contribution. Correspondence to ha.li@campus.lmu.de}\def\thefootnote{\arabic{footnote}}

\ificcvfinal\thispagestyle{empty}\fi

\begin{abstract}

The field of multimodal research focusing on the comprehension and creation of both images and text has witnessed significant strides. This progress is exemplified by the emergence of sophisticated models dedicated to image captioning at scale, such as the notable Flamingo model and text-to-image generative models, with DALL-E serving as a prominent example. An interesting question worth exploring in this domain is whether Flamingo and DALL-E understand each other. To study this question, we propose a reconstruction task where Flamingo generates a description for a given image and DALL-E uses this description as input to synthesize a new image. We argue that these models understand each other if the generated image is similar to the given image. Specifically, we study the relationship between the quality of the image reconstruction and that of the text generation. We find that an optimal description of an image is one that gives rise to a generated image similar to the original one. The finding motivates us to propose a unified framework to finetune the text-to-image and image-to-text models. Concretely, the reconstruction part forms a regularization loss to guide the tuning of the models. Extensive experiments on multiple datasets with different image captioning and image generation models validate our findings and demonstrate the effectiveness of our proposed unified framework. As DALL-E and Flamingo are not publicly available, we use Stable Diffusion and BLIP in the remaining work. Project website: \url{https://dalleflamingo.github.io}.
\end{abstract}

\begin{figure}[t]
    \centering
    \includegraphics[width=\linewidth]{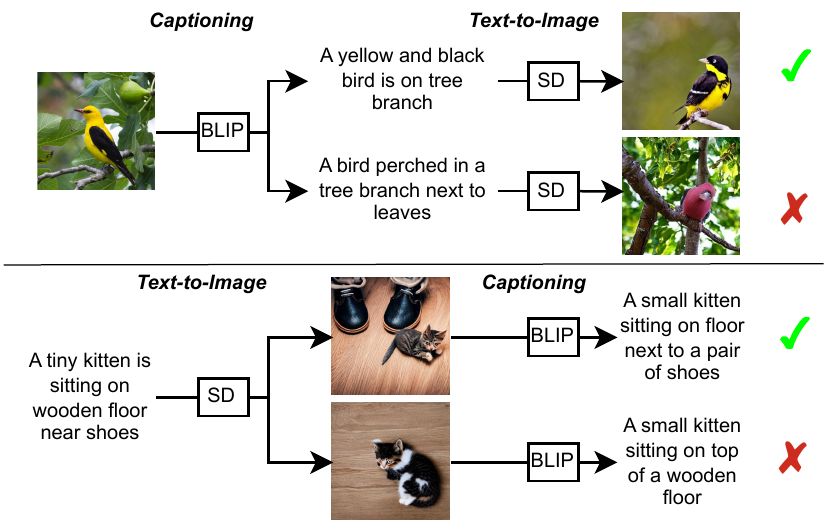}
    \caption{Illustration of the communication tasks between SD and BLIP. Top: SD generates an image for each caption created by BLIP, where an accurately reconstructed image indicates a more precise caption. Bottom: The reverse task involves SD generating image candidates, which are then used by BLIP to produce captions for those images. The best image that represents a text is the one that leads to the best reconstruction of the original text.}
    \label{fig:intro}
\end{figure}

\section{Introduction}

Recently, multimodal research that aims to improve machine understanding of images and text has made significant advances~\cite{radford2021learning, ramesh2022hierarchical, rombach2022high, bayoudh2021survey, lu2022unified, yasunaga2022retrieval,zou2023generalized,gu2023systematic}. Text-to-image generation models such as DALL-E~\cite{ramesh2021zero,ramesh2022hierarchical} and Stable Diffusion (SD)~\cite{rombach2022high} are capable of converting complex textual descriptions~\cite{saharia2022photorealistic} from real-world scenarios into high-fidelity images~\cite{ramesh2022hierarchical, ho2020denoising,nichol2021glide, ramesh2021zero}. Conversely, image-to-text generation models, e.g., Flamingo~\cite{alayrac2022flamingo} and BLIP~\cite{li2022blip}, exhibit the ability to comprehend the intricate semantics present in images and produce coherent descriptions~\cite{xu2015show,hossain2019comprehensive,wang2021simvlm, wang2022unifying, zhang2021vinvl, li2020oscar, yu2022coca}. Despite the closeness of the image captioning and text-to-image generation tasks, they are often studied in isolation from each other, i.e., the communication between these models is under-explored~\cite{lu2022unified, kim2022verse}. 
This raises an interesting question: do image-to-text generation models and text-to-image generation models possess mutual understanding? Concretely, we investigate this question by letting an image-to-text model, BLIP, generate a text description for a given image, which subsequently serves as input to a text-to-image model, SD, to synthesize a new image\footnote{The initial idea of this work was motivated by Flamingo and DALL-E. However, their model weights are unavailable at the time of publication.}. We argue that BLIP and SD understand each other if the generated image is similar to the source image. Such mutual understanding may enhance their respective abilities to comprehend underlying concepts, resulting in superior caption generation and image synthesis. Figure \ref{fig:intro} illustrates this idea, where the upper caption is a better representation of the input image than the lower caption as it leads to a more faithful reconstruction of the original image.

To verify this assumption, we design two reconstruction tasks: image-text-image and text-image-text, shown in Figure \ref{fig:intro}. For the first reconstruction task, we evaluate the similarity between the semantics of the generated image and the input image, e.g., by computing the distance of image features extracted with a pretrained CLIP image encoder~\cite{radford2021learning}. Afterward, we compare the generated text with human-annotated captions to assess the quality of the generated text~\cite{vedantam2015cider}. Our experiments reveal that the quality of the reconstruction depends on the quality of the generated text. This leads to our first finding: the best description for an image is the description that enables the generative model to recreate the original image. Similarly, we design the reverse task where SD generates an image from a given text, and subsequently, BLIP produces a text from the generated image. We find that the best image representation for text is the one that generated the original text. We conjecture that through the reconstruction task, information on the input image is well preserved in the textual description and that meaningful description leads to a faithful recovery back to the image modality.

Based on our findings, we propose a novel finetuning framework that facilitates communication between text-to-image and image-to-text models, enabling them to talk to each other. Specifically, in our framework, a generative model not only receives training signals from human labels but also from a reconstruction loss. For a given image or text, one model first generates a representation of the input in the other modality and then the other model converts this representation back to the input modality. The reconstruction part forms a regularization loss to guide the finetuning of the first model. In this way, they acquire not only human supervision but also self-supervision that the generation should lead to a more accurate reconstruction. For example, the image captioning model should favor captions that not only match the labeled image-text pairs but also those that can lead to reliable reconstructions.

Our work is closely related to inter-agent communication. Language is a major means of exchanging information between agents. But how can we be sure that the first agent has the same understanding of what a cat or a dog is as the second agent? In this paper, we have the first agent analyze an image and produce a text describing that image. The second agent then obtains the text and simulates an image based on the text. This latter process can be thought of as an embodiment process~\cite{tresp2023tensor}. We propose that communication is successful if the image simulated by the second agent is close to the image the first agent received as input. In essence, we test the effectiveness of language, which is the main communication venue of humans.

We conduct experiments leveraging the off-the-shelf  models~\cite{radford2021learning,li2022blip,caron2021emerging,reimers2019sentence,rombach2022high,zhou2021lafite}, in particular recently developed large-scale pre-trained image captioning models~\cite{li2022blip,li2023blip} and image generation models~\cite{rombach2022high,zhou2021lafite}. Extensive experiments demonstrated the advantages of our proposed framework for various generative models, in both training-free and finetuning settings. Specifically, in the training-free paradigm, our framework significantly enhanced the caption and image generation, whereas, for finetuning, we achieved improved results for both generative models. Our main contributions are summarized as follows:
\begin{itemize}
    \item \textbf{Framework:} To the best of our knowledge, we are the first to explore the communication of standard alone image-to-text and text-to-image generative models through human-interpretable text and image representations. Contrastively, related work unifies image and text generation implicitly through an embedding space.
    \item \textbf{Findings}: We find that the quality of a caption can be evaluated by assessing the image reconstruction produced by a text-to-image model. The best caption for an image is one that leads to the most accurate reconstruction of the original image. Similarly, the best image for a caption is the image that leads to the best reconstruction of the original text.
    \item \textbf{Improvements}: Based on our findings, we propose a unified framework to enhance both the image-to-text and text-to-image models. This involves finetuning the image-to-text model using a reconstruction loss computed by a text-to-image model as regularization, and finetuning the text-to-image model using a reconstruction loss computed by an image-to-text model. We analyzed and verified the effectiveness of our framework.
\end{itemize}
\section{Related Work}
\vspace{0.1cm}
\noindent \textbf{Text to Image Generation}
Popular text-conditioned image generation models mainly include GAN~\cite{goodfellow2020generative, karras2020analyzing, karras2019style}, VAE~\cite{kingma2014auto, zhou2021lafite, ding2021cogview, gafni2022make, yu2022scaling}, and recently developed diffusion models~\cite{ho2020denoising, nichol2021glide, rombach2022high, ramesh2022hierarchical,saharia2022photorealistic, balaji2022ediffi}. Diffusion models model image generation as a Markov Chain and learn the reversed process, where a noise vector is gradually denoised into an image~\cite{ho2020denoising}. For text-guided image generation, generative models compute the conditional probability of generating an image given the text. DALL-E~\cite{ramesh2022hierarchical} and SD~\cite{rombach2022high} are representatives of such diffusion-based models that are scaled to real-world complexity. Such large-scale generative models are trained on an extensive amount of image-text pairs like the LAION~\cite{schuhmann2021laion} dataset obtained from the web. The utilization of large-scale datasets allows these models to generate a vast diversity of images from intricate text inputs. The focus of this work is specifically on examining the mutual understanding of these large-scale models.

\begin{figure*}[htp]
    \centering
    \includegraphics[width=0.95\textwidth]{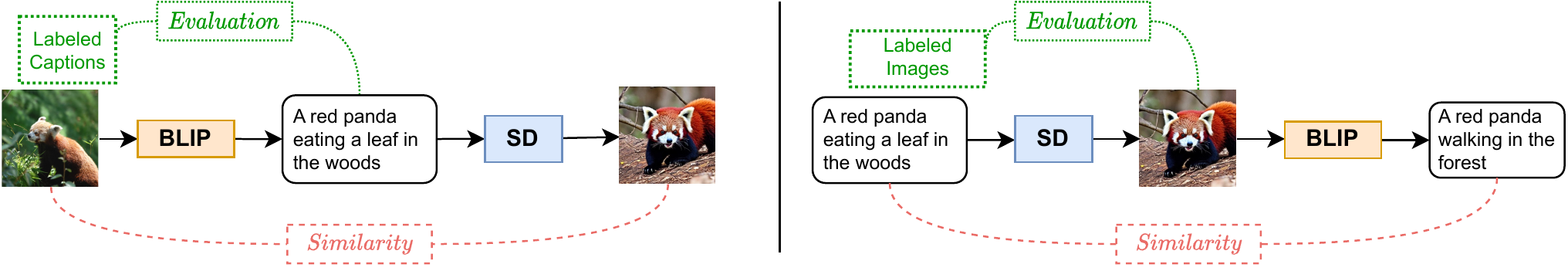}
    \caption{Illustration of our proposed inference framework. Left: a pipeline for image-text-image. The input image is fed to BLIP for caption generation and SD reconstructs the image from the generated text. The generated image is compared with the input image using a similarity function, e.g., based on image embeddings, which is utilized to evaluate the quality of the generated caption. We treat human-annotated captions as ground truth representations of the input image. Right: a pipeline for text-image-text. The reverse task for reconstructing text is demonstrated here.
    } 
    \label{fig:architecture}\vspace{-0.18cm}
\end{figure*}

\vspace{0.1cm}
\noindent \textbf{Image Captioning}
Image captioning describes a scene using natural language~\cite{hossain2019comprehensive,wang2021simvlm, wang2022unifying, zhang2021vinvl, li2020oscar, yu2022coca}. As one of the representatives of image captioning models, BLIP consists of an image encoder to understand the image features and a text decoder to generate text in an autoregressive manner. The image encoder uses a vision-transformer~\cite{dosovitskiy2020image} backbone, which divides an image into a sequence of patches and outputs a sequence of embeddings that serve as the grounding information for text generation. Thereafter, a text decoder predicts the next token by attending to previous tokens and the encoded visual states. A recent trend in image captioning research is to develop large-scale visual language models that unify text generation with multiple image-text understanding tasks, e.g., image-based question answering and image-text retrieval~\cite{alayrac2022flamingo,cho2021unifying, yu2022coca}. We utilize BLIP as the image captioning method in this work.

\vspace{0.1cm}
\noindent \textbf{Vision Language Representation Learning and Understanding}
Representation learning for vision and language handles semantic alignment between different modalities~\cite{tan2019lxmert, chen2019uniter,cho2021unifying,zhu2022uni,bao2021vlmo,gu2023systematic}. A popular approach is to utilize contrastive learning on large-scale image and text pair datasets to obtain a unified representation of different representations for the same concept~\cite{radford2021learning}. For unimodal image representation, self-supervised visual encoders, e.g., DINO~\cite{caron2021emerging}, are proposed to encode visual semantics. For unimodal text representation, language models such as SBERT~\cite{reimers2019sentence} can well extract semantics from the text. 
Several works have been proposed to enhance image and text generation by leveraging aligned image and text representations in the embedding space~\cite{lu2022unified,bao2023one,xu2022versatile}. In contrast to their work, we aim to understand the representations learned by different image-to-text and text-to-image generative models using text or image as an interface.

\section{Findings: Do BLIP and Stable Diffusion Understand Each Other?}
\label{sec:finding}

In this section, we introduce two reconstruction tasks to test the mutual understanding of frozen BLIP and SD. We present our findings that a training-free reconstruction process can enhance the quality of a generated text or image.

\subsection{Task Setup}
\label{sec:finding_setup}

\vspace{0.1cm}
\noindent \textbf{Image-Text-Image}
This task aims to study the relationship between the quality of a reconstructed image and that of a caption. As shown on the left side of Figure \ref{fig:architecture}, BLIP describes a given image in text, and SD generates an image from this description. Firstly, BLIP describes a source image $x$ with different caption candidates $\{y^{(i)}\}_{i=1:N}$. When generating the captions, we use Top-$p$ sampling~\cite{holtzman2019curious} as the default sampling method while also examining two additional sampling strategies. Secondly, the captions are given to the SD to generate one image ${\hat{x}}^{(i)}$ for each caption $y^{(i)}$. We acquire $N$ generated images for each input image. Finally, we compute the similarity between a generated image and its corresponding input image, as shown in red dashed lines in Figure \ref{fig:architecture}. 
We utilize image embeddings obtained from pretrained encoders such as CLIP image encoder and DINO. The image encoder first encodes the input image $x$ and a generated image $\hat{x}^{(i)}$ into embeddings space and the cosine similarity between the two embeddings is computed. We also experiment with additional image fidelity measures for a more solid evaluation.

\vspace{0.1cm}
\noindent \textbf{Text-Image-Text}
Likewise, in the setting of \textit{SD talking to BLIP}, shown on the right side of Figure \ref{fig:architecture}, we investigate whether an improved text reconstruction corresponds to a superior image representation for a given text. We randomly sample a text from the image-text pair dataset and generate $N$ images for each text using SD. Then BLIP generates a description for each input image using beam search~\cite{freitag2017beam}. Following this, we use different methods to calculate the similarity between the input and generated text, which is shown as the red dashed line in Figure \ref{fig:architecture}. The similarity is computed with a text encoder where the alignment between two text embeddings extracted by the encoder is calculated. Two encoders are applied in our work. The first one is a multimodal text encoder, i.e., CLIP text encoder trained to align text and image features; The second one is a unimodal text encoder, i.e., SBERT~\cite{reimers2019sentence} trained only on text data. Besides, traditional text similarity metrics like CIDEr~\cite{vedantam2015cider} and WMD~\cite{pmlr-v37-kusnerb15} between texts are also applied.

\subsection{Evaluation Protocols}
\label{sec:finding_eval}
\vspace{0.1cm}
\noindent \textbf{Evaluation Metric}
\textit{Image Captioning:} We report standard metrics, including BLEU~\cite{papineni2002bleu}, CIDEr~\cite{vedantam2015cider}, SPICE~\cite{anderson2016spice}, and WMD~\cite{pmlr-v37-kusnerb15}, as well as an embedding-based metric that computes the cosine distance between the embeddings of a candidate text and a human-labeled reference caption. For that, the CLIP text encoder and SBERT text encoder are used to obtain the caption embeddings.
\textit{Image Generation:} We use standard fidelity scores like FID~\cite{heusel2017gans} and Inception Score (IS)~\cite{salimans2016improved} to quantify the quality of the generated images. Similar to the caption evaluation, we also report embedding-based image distances between a generated and a real reference image. For that, we report the CLIP Visual Score similar to \cite{kumari2022multi} and the DINO Score~\cite{caron2021emerging}.

\vspace{0.1cm}
\noindent \textbf{Implementation Details} 
We set the number of caption candidates to $N$=10. The NoCaps~\cite{agrawal2019nocaps} validation set and the COCO Karpathy test split~\cite{karpathy2015deep} are used to support the evaluation. We randomly sample a caption for each image in the dataset for the input of the text-image-text task. All models are frozen in this section, without any finetuning. More implementation details can be found in Appendix \ref{app:finding_exps}.

\subsection{Findings}

\begin{figure}[htp]
    \centering
    \includegraphics[width=\linewidth]{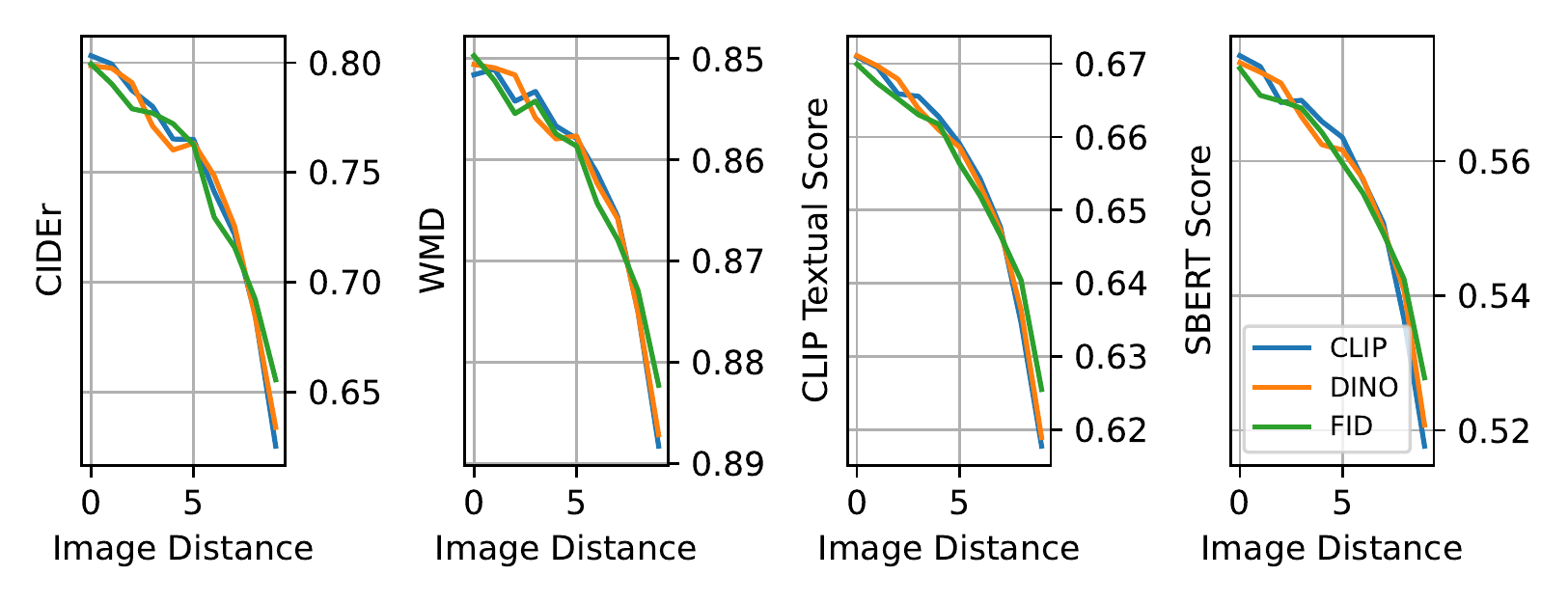}
    \caption{Evaluation of the image-text-image pipeline with three similarity and four caption metrics on the NoCaps dataset. The result suggests that regardless of the similarity or evaluation metric, better image reconstruction always leads to better captions.
    }
    \label{fig:finding_image}
\end{figure}

\noindent \textbf{Insight I: a Better Caption is the One That Leads to a Better Visual Reconstruction.}
In the setting of image-text-image, we compute the captioning score of each generated text description, as well as the similarity score between the corresponding generated image and the input image. Then we rank the image similarities within $N$ generated pairs for each input image and then aggregate the caption scores over the entire dataset. Figure \ref{fig:finding_image} displays the correlation between the quality of the image reconstruction and the caption quality. Four scoring metrics and three similarity methods are analyzed in this experiment.

It is evident from Figure \ref{fig:finding_image} that the reconstruction quality evaluated by a text-to-image model reveals the quality of the generated caption for the input image. Specifically, the better the reconstructed image, the better the caption score, independent of the similarity and the evaluation metrics. Moreover, DINO shows on-par performance with CLIP, even though DINO is trained only with image augmentations and contrastive learning on the pixel space. This rules out the possibility that CLIP might influence the comparison since it is trained to align multimodal image-text features. Furthermore, we obtain consistent results for both embedding-based captioning metrics and word frequency matching metrics.

Quantitatively, we compare our method with the baseline sampling method. For baseline, we repeat the sampling multiple times and average the final captioning metric. As shown in Table \ref{tab:finding1}, our approach consistently produces better captions. The relative gain is more significant for NoCaps, especially for the out-domain subset, which contains novel objects that usually require accurate concepts to describe. The choice of the hyperparameter for the baseline sampling method is discussed in Appendix \ref{app:top_p}. Additionally, we find our conclusion consistent for two different sampling strategies widely used for captioning (See Appendix \ref{app:sampling_methods}). The number of candidates $N$ has minimal impact on the conclusion and is discussed in Appendix \ref{app:finding_num}.

Qualitatively, we observed plausible improvements. Our approach accurately describes the bird in Figure \ref{fig:qualitative} with detailed attributes. More figures are in Appendix \ref{app:finding_qualitative}.

\begin{figure}[htp]
    \centering
    \includegraphics[width=\linewidth]{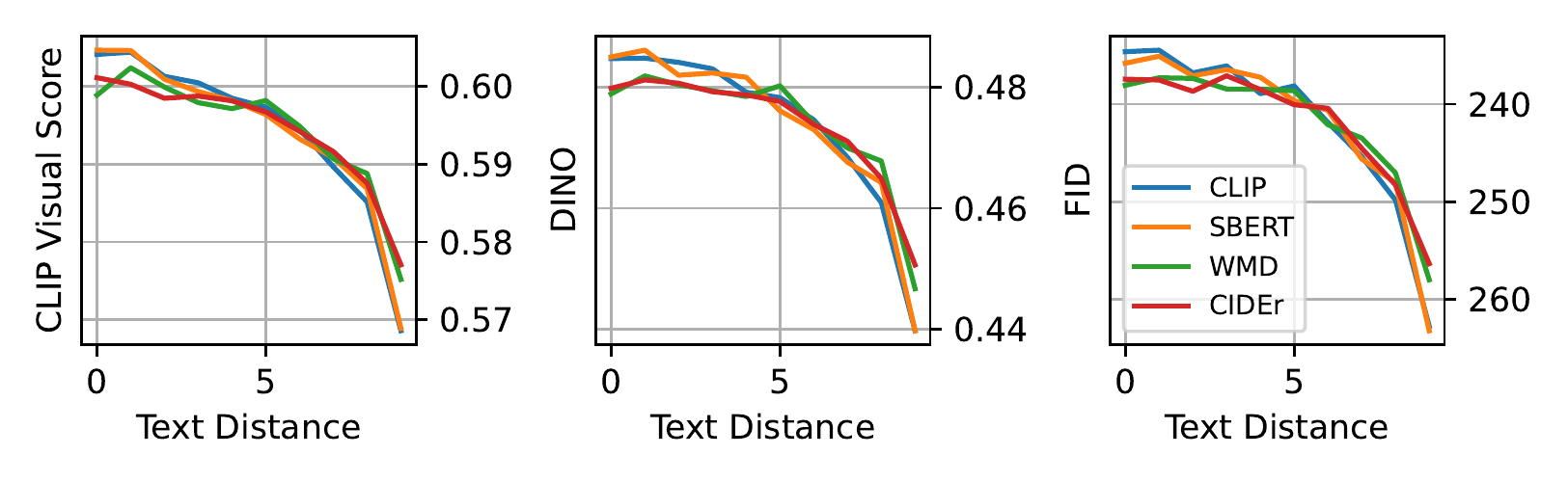}
    \caption{Evaluation of the text-image-text pipeline with four types of similarity and three types of evaluation metrics on the NoCaps dataset. The result is consistent for all types of combinations.}
    \label{fig:finding_text}
\end{figure}

\begin{table*}
    \centering
    \resizebox{\textwidth}{!}{
    \scriptsize
    \begin{tabular}{l|cccccccc|cccc}\toprule
    & \multicolumn{8}{c}{Nocaps}& \multicolumn{4}{|c}{COCO} \\
    \cline{2-13}
    & \multicolumn{2}{c}{In-domain}& \multicolumn{2}{c}{Near-domain}& \multicolumn{2}{c}{Out-domain} & \multicolumn{2}{c}{Overall}& \multicolumn{4}{|c}{Karpathy Test} \\
    Method& CIDEr & SPICE  & CIDEr  & SPICE  & CIDEr  & SPICE  & CIDEr  & SPICE &B@3 & B@4 & CIDEr & SPICE \\\hline
Baseline Sampling & 75.1 & 12.4 & 72.4 & 11.9 & 78.7 & 11.5 & 74.1 & 11.9 &32.4 & 21.9 & 90.1 & 19.6\\
Ours &\textbf{77.3} & \textbf{12.9} & \textbf{78.3} & \textbf{12.6} & \textbf{88.8} & \textbf{12.4} & \textbf{80.3} & \textbf{12.6} &  \textbf{32.5}& \textbf{22.0} & \textbf{92.0} & \textbf{20.1}\\ \hline
Gain (\%) & +2.9 & +4.0 & +8.1 & +5.9 & +12.8 & +7.8 & +8.4 & +5.9 & +0.4 & +0.3 &+2.1 & +2.2\\\toprule
    \end{tabular}
    }
    \caption{Comparison of the baseline captioning model and our proposed method on Nocaps and COCO datasets. Our method outperforms the baseline sampling method on all metrics. The relative gain of our method compared to the sampling method is given in the third row. B@$k$: BLEU@$k$.}
    \label{tab:finding1}
\end{table*}

\begin{table}[htp]
    \centering
    \resizebox{\linewidth}{!}{
    \begin{tabular}{l|ccc|ccc}\toprule
    & \multicolumn{3}{c}{NoCaps} & \multicolumn{3}{|c}{COCO} \\
    Model &  CLIP$\downarrow$ &FID$\downarrow$ &IS$\uparrow$ & CLIP$\downarrow$  &FID$\downarrow$  &IS$\uparrow$  \\\hline
    Sampling & 40.54 & 32.37  & 41.19 & 42.54 & 44.76 & 30.01\\
    Ours  & \textbf{33.47} & \textbf{29.59} & \textbf{45.64} & \textbf{34.74} & \textbf{42.02} & \textbf{31.62} \\ \hline
    Gain (\%)  & +17.4 & +8.6 & +9.8 & +18.3 & +6.1 & +5.4 \\
    \toprule
    \end{tabular}
    }
    \caption{Comparison of our proposed method to SD on image generation. Our method uses BLIP to filter out generated images based on text reconstruction.}
    \label{tab:finding_text}
\end{table}

\begin{figure*}[htp]
    \centering
    \includegraphics[width=\linewidth]{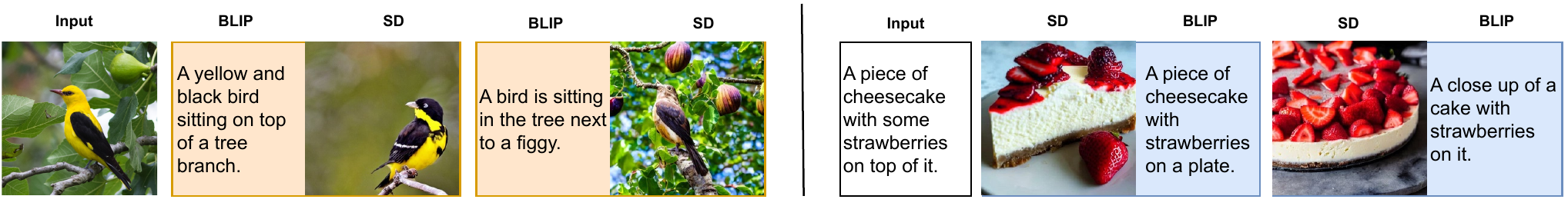}
    \caption{Qualitative examples of reconstruction tasks. The left side shows image reconstruction for the given input image of a bird. Two samples are shown with their generated images, ranked by the image similarity. The right side shows text reconstruction whereas the first sample gives a high-quality image as well as high-quality text reconstruction.}
    \label{fig:qualitative}
\end{figure*}

\vspace{0.1cm}
\noindent \textbf{Insight II: a Better Image is the One That Leads to a Better Text Reconstruction.}
Similarly, we find that text reconstruction boosts the quality of text-to-image generation. As shown in Figure \ref{fig:finding_text}, when the reconstructed text is dissimilar to the original text, on average the image quality is low. In contrast, high-quality reconstruction is associated with a high-quality image. This holds true for both the fidelity and semantic metrics on image generation evaluation. For different similarity functions, we find that embedding-based similarity methods, i.e., CLIP and SBERT, perform better than similarity metrics based on word co-occurrence, i.e., CIDEr and WMD. Table \ref{tab:finding_text} demonstrates the quantitative improvement of our method compared to the baseline in terms of fidelity and semantic alignment. Additionally, we present a qualitative example in Figure \ref{fig:qualitative}. The last image does not properly represent the input text describing a piece of cheesecake, which leads to an inaccurate caption in the next step. By comparing the generated captions with the input text, we find better images that faithfully depict the text. These examples highlight the finding that an image-to-text model can be used to assess the quality of a generated image, i.e., the more similar the reconstructed text to the input text, the higher the quality of the generated image.

Additional experiments with different image captioning and text-to-image generation models in Appendix \ref{app:diff_models} lead to the same conclusion for the two tasks.

\begin{figure*}[htp]
    \centering
    \includegraphics[width=\linewidth]{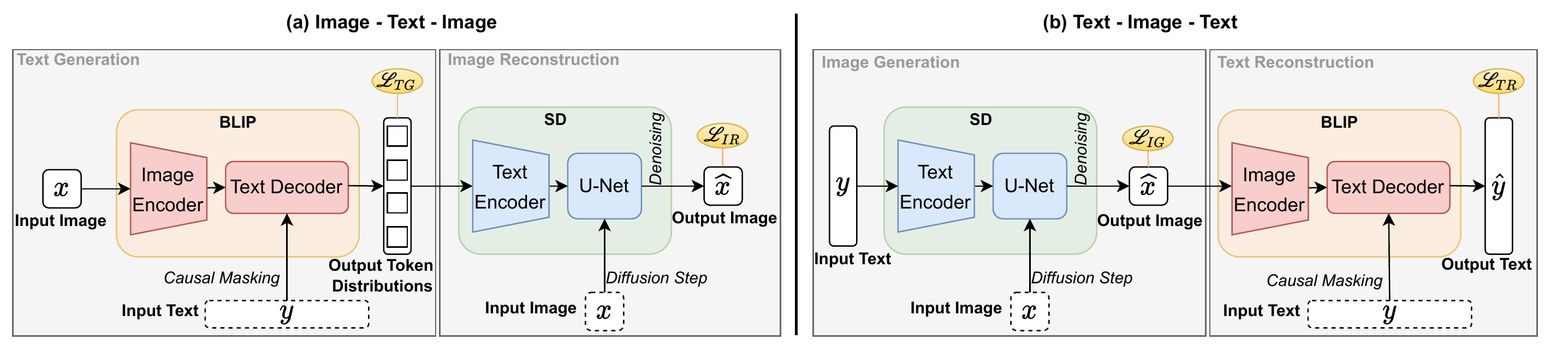}
    \caption{Illustration of our proposed finetuning framework. We introduce a reconstruction pipeline starting from an image (left) and a reconstruction pipeline starting from the text (right). Input images and text shown at the bottom (dashed boxes) are only used during training and will be dropped during inference. For illustration proposes, we omit the VAE in SD without changing the principles. Causal masking is a technique for language modeling during training. The diffusion step denotes adding noise to the input image.}
    \label{fig:loss}
\end{figure*}

\section{Method: Let BLIP and Stable Diffusion Talk}
Based on the insights in the last section, we introduce a novel approach to finetuning the image captioning and text-to-image models by incorporating reconstructions as regularization losses. Similar to the last section, we introduce two pipelines: image-text-image and text-image-text. In the first pipeline, BLIP generates a caption for a given image, where SD is guided by this caption to reconstruct the input image. The core component is a differentiable layer that connects the output of the BLIP with the input of SD. This connection allows optimizing BLIP using the loss computed by SD in the text-to-image generation process. Likewise, in the second pipeline, we optimize the SD model using the loss obtained from comparing the generated text by BLIP with the input text. A schematic of our training framework is presented in Figure \ref{fig:loss} and explained below.

\subsection{Image-Text-Image}
\label{pipeline1}
\vspace{0.1cm}\noindent \textbf{Text Generation Stage} 
The left box in Figure \ref{fig:loss} (a) illustrates the standard image captioning process using the BLIP model. BLIP takes an image $x \in \mathbb{R}^{H\times W \times 3}$ in RGB space as input and produces a sequence of tokens $y_t$. The prediction of each token $y_t$ at time step $t$ relies on tokens generated in previous steps $y_{<t}$ and the image embeddings. To accelerate the training using batch operation, a ground-truth caption $y$ is fed to BLIP with an attention mask, ensuring that each token's prediction is causally dependent on the tokens that came before it. In this way, BLIP can generate a caption for an input image using only a single forward pass during training.
Therefore, the training objective of the image captioning model is to minimize the cross-entropy loss between the ground truth text and the predicted text, defined as
\begin{equation}
    \mathcal{L}_{TG} = \mathbb{E}_{x,y \sim \mathcal{D}} \big[ \Pi_t p_{\theta}(y_{t}|y_{<t},x) \big],
\end{equation}
where $\theta$ refers to the weights of BLIP and $\mathcal{D}$ refers to the dataset from which a ground-truth image-text pair $(x,y)$ is sampled. The BLIP model was pretrained using $\mathcal{L}_{TG}$. Throughout the finetuning process, this loss is further utilized to update the weights of BLIP, preventing its predictions from deviating from the text used in pretraining.

\vspace{0.1cm}\noindent \textbf{Differentiable Connection}
The final softmax layer of BLIP generates a token distribution $\hat{g}_t \in \mathbb{R}^V$ at each timestep which is used to decode a discrete token $y_t$ during image captioning. $V$ represents the size of the vocabulary of a tokenizer. However, instead of sampling a specific $y_t$ at each step, we store the token distributions $\hat{g} \in \mathbb{R}^{L \times V}$ for all timesteps, where $L$ is the number of tokens in a caption. During the training phase, this can be obtained after a forward process of the input image. Subsequently, we compute the dot product of the token distributions $\hat{g}$ with the vocabulary embeddings $E\in \mathbb{R}^{V\times D}$ of SD's tokenizer. This produces text embeddings, which are used to guide image generation. Later stages follow the standard text-to-image generation pipeline.

In addition, to address the discrepancy between the tokenizers used by BLIP and SD, we employ a hard-coded transformation matrix to map the vocabulary of BLIP's tokenizer to that of SD (See Appendix \ref{app:training_tokenizer}). For future work, we will explore discrete sampling methods for caption generation. In this work, we stick to our simple strategy as it has proven to be useful in practice.

\vspace{0.1cm}\noindent \textbf{Image Reconstruction Stage}
The right box of Figure \ref{fig:loss} (a) shows the conventional procedure of training a text-guided image generative diffusion model. The model is trained on an input image $x$ along with its textual description $y$. The training procedure involves a forward diffusion process wherein a clean image is progressively destructed by introducing noise in iterative steps. Then SD learns a reversed process where it predicts the noise that is necessary to reconstruct the denoised image for each step. In the context of text-to-image generation, the input text $y$ is processed by the text encoder as conditional information to guide this reversed denoising process. 

Specifically, for the diffusion process, a noise vector $\epsilon$ for timestep $t$ is sampled from a normal distribution. A noised image $x_t$ of the original clean image $x_0$ for timestep $t$ is computed following a predefined noise scheduling process denoted as $x_t= \sqrt{\alpha_t} x_0 + \sqrt{1-\alpha_t} \epsilon$, where $\alpha_t$ is a predefined scalar value for noise scheduling. The U-Net
\footnote{In fact, SD utilizes a variational autoencoder to compress the image into a latent space and subsequently learns the diffusion process in that space. For more details, readers are referred to~\cite{rombach2022high}.} of SD predicts $\epsilon$ from $x_t$, $t$, and the encoding of the text input $c$. Formally, the training objective of SD is to minimize a mean squared error (MSE), defined as
$\mathcal{L} = \mathbb{E}_{x,y \sim \mathcal{D}, t \sim [1,T]} \Big[\Vert \hat{\epsilon}_{\psi}(x_t, t, c) - \epsilon \Vert^2 \Big],$
where $\psi$ represents the weights of SD and $T$ is a fixed number for diffusion steps. SD uses a CLIP text encoder $\pi$ to encode the text $y$, written as $c=\pi(y)$. In our proposed approach, the encoding is obtained using $c=\pi(\hat{g})$, which incorporates the modification introduced in the preceding subsection.
In summary, the image reconstruction loss is,
\begin{equation}
\label{eq:l_ir}
    \mathcal{L}_{IR} =  \mathbb{E}_{x,y \sim \mathcal{D}, t \sim [1,T]} \Big[\Vert \hat{\epsilon}_{\psi}(x, t, \pi(\hat{g})) - \epsilon \Vert^2 \Big].
\end{equation}
Intuitively, if a caption is of high quality in describing the content of the input image, SD is expected to have a lower loss in reconstructing that image. During training, we uniformly sample a timestep $t$ for each predicted caption.

In contrast to the approach outlined in Section \ref{sec:finding_setup}, our novel design does not require an extra image encoder since the input image is already incorporated into the generation process of SD. This aligns with the standard training procedure, where SD reconstructs a given ground-truth image rather than generating a new image from scratch. Directly converting the approach in Section \ref{sec:finding_setup} into a training framework would be impractical as it requires a costly sampling procedure.
 
\subsection{Text-Image-Text}
\label{pipeline2}

\vspace{0.1cm}\noindent \textbf{Image Generation Stage}
This stage is equivalent to the conventional training of SD described above. Similar to Eq. \ref{eq:l_ir}, the loss is defined as
\begin{equation}
\label{eq:l_ig}
\mathcal{L}_{IG} = \mathbb{E}_{x,y \sim \mathcal{D}, t \sim [1,T]} \Big[\Vert \hat{\epsilon}_{\psi}(x_t, t, \pi(y)) - \epsilon \Vert^2 \Big].
\end{equation}
However, this procedure does not directly produce a clean image, which is required as input for BLIP. For that, we adopt the 1-step approximation~\cite{li2022upainting, wallace2023end} technique for diffusion models. Specifically, based on the noisy image $x_t$ and SD's predicted noise $\hat{\epsilon} = \hat{\epsilon}_{\psi}(x_t, t, \pi(y))$, a clean image $\hat{x}_0$ can be recovered by $\hat{x}_0 = \frac{1}{\sqrt{\alpha_t}} (x_t - \sqrt{1-\alpha_t} \hat{\epsilon})$. During training, for each sampled timestep $t$, we predict $\hat{x}_0$ and feed the prediction to BLIP for caption generation.

\begin{table*}[htp]
    \centering
    \small
    \resizebox{\textwidth}{!}{
    \begin{tabular}{l|ccccccccc|cccc}\toprule
    & \multicolumn{9}{c|}{Nocaps} & \multicolumn{4}{c}{COCO} \\
    Method &I-C & N-C & O-C & B@1 & B@2 & B@3 & B@4 & CIDEr & SPICE & B@3 & B@4  & CIDEr & SPICE  \\\hline
    BLIP ViT-B~\cite{li2022blip}  & 111.8 & 108.6 & 111.5 & 83.6 & 68.2 & 50.6 & 32.0 & 109.7 & 14.7 & 50.5 & 39.7 & 133.3 & 23.8 \\
    Ours ViT-B, SD& \textbf{114.9} & \textbf{110.8} & \textbf{112.9} & \textbf{84.6} & \textbf{69.4} & \textbf{51.9} & \textbf{32.9} & \textbf{111.8} & \textbf{14.9}
& \textbf{51.4} & \textbf{40.1} & \textbf{134.6} & \textbf{24.0} \\
    \hline\hline
    BLIP ViT-L~\cite{li2022blip} & 114.9 & 112.1 & 115.3 & 84.2 & 69.3 & 51.7 & 33.1 & 113.2 & 14.8 &  51.4 & 40.4 & 136.7 & \textbf{24.3} \\
    Ours ViT-L, SD& \textbf{116.1} & \textbf{113.0} & \textbf{115.6} & \textbf{84.7} & \textbf{70.0} & \textbf{52.5} & \textbf{33.6} & \textbf{114.0} & \textbf{14.9} & \textbf{52.0} & \textbf{40.9} & \textbf{137.3} & 24.1\\ \hline\hline
    BLIP-2~\cite{li2023blip} & 122.7 & 118.0 & 123.8 & 86.8 & 73.2 & 56.2 & 37.0 & 119.9 & \textbf{15.4} & 55.5 & 43.7 & 145.8 & \textbf{25.2} \\
    Ours BLIP-2, SD & \textbf{123.8} & \textbf{119.3} & \textbf{124.2} & \textbf{86.9} & \textbf{73.5} & \textbf{56.5} & \textbf{37.1} & \textbf{121.0} & \textbf{15.4} &\textbf{56.0} & \textbf{44.1} & \textbf{146.4} & \textbf{25.2} \\
    \toprule
    \end{tabular}
    }\vspace{0.3cm}
    \caption{Evaluation results of image captioning. We conduct experiments with different image captioning models (BLIP ViT-B, BLIP ViT-L, BLIP-2) with an image generation model (SD). I-C/N-C/O-C: In-/Near-/Out-domain CIDEr. The remaining metrics are on the entire set. B@$k$: BLEU@$k$.}
    \label{tab:image}
\end{table*}

\vspace{0.1cm}\noindent \textbf{Text Reconstruction Stage}
The BLIP takes the predicted image $\hat{x}_0$ as input and produces a caption. Following the same procedure in the text generation stage, the text reconstruction loss is defined as
\begin{equation}
    \mathcal{L}_{TR} = \mathbb{E}_{x,y \sim \mathcal{D}} \big[ \Pi_t p_{\theta}(y_{t}|y_{<t},\hat{x}_0)\big].
\end{equation}
The major difference is that $\mathcal{L}_{TR}$ is conditioned on images with gradients originating from SD, which allows optimizing SD's parameters by BLIP's loss.

\subsection{Full Training Objective}
We simplify the training pipelines by adding the individual losses into a single training loss, and then optimizing both models on this summed loss. The parameters of both models are updated at each iteration, enabling a joint improvement of both models.
\begin{equation}
\label{eq:loss}
    \mathcal{L}(\theta, \psi) = \mathcal{L}_{TG} + \mathcal{L}_{IR} + \mathcal{L}_{IG} + \mathcal{L}_{TR}.
\end{equation}
See Appendix \ref{app:pseudo_code} for a discussion on our loss function, including its connection to CycleGAN~\cite{zhu2017unpaired}, and the pseudo-code for the training framework.

\section{Experimental Setup}

\vspace{0.1cm}
\noindent \textbf{Dataset and Evaluation} \textit{Training dataset:}
We finetune both BLIP and SD on the COCO Karpathy train split~\cite{karpathy2015deep} of 113k images, each associated with five captions. \textit{Image Captioning:} Following~\cite{li2022blip}, the image captioning is evaluated on the COCO test set and NoCaps validation set, utilizing metrics described in Section \ref{sec:finding_eval}. Unlike the COCO dataset, which contains images with common object categories, the NoCaps dataset includes images in the wild, making it a challenging benchmark for zero-shot evaluation for image captioning. \textit{Image Generation:} For image generation, we report the CLIP image distance, defined in Section \ref{sec:finding_eval}, on the COCO test set and the NoCaps validation set. Likewise, the NoCaps dataset is used as a zero-shot evaluation benchmark. As each image is associated with multiple captions, we randomly sample a caption for each image to construct these two test sets.

\vspace{0.1cm}
\noindent \textbf{Baselines and Models} \textit{Image Captioning:} BLIP ViT-B denotes the BLIP model with ViT-B/16 as the visual backbone. BLIP ViT-L uses a larger variant of ViT, i.e., ViT-L/16. We use the bootstrapped and finetuned version since they are optimized to produce the best baselines for captioning. We further explore the BLIP-2 ViT-G OPT$_{\mathrm{2.7B}}$ which is among the SOTA captioning methods. We do not further finetune their models as they have already been finetuned on the same dataset using text generation loss.
Thus we use performance metrics reported from their paper~\cite{li2022blip,li2023blip}. For the metrics that are not available, we run the evaluation with their published code and weights to get the results. \textit{Image Generation:} Since, as of this writing, we have not found any other publicly available text-to-image generation models with on-par performance, we utilize SD as the baseline model. We use the weights of sd-v1-4.ckpt for SD. The output image size is 512 except for BLIP-2, where we downscaled the image size to 384 due to hardware constraints. Since SD is not trained on the COCO dataset, we finetune SD using MSE loss to serve as a baseline method.

For our finetuning framework, we conducted experiments with the combinations of the three above-mentioned image captioning models and the text-to-image model. For the ablation study, we use BLIP ViT-B as the default setting.

\vspace{0.1cm}
\noindent \textbf{Training Details}
To improve the efficiency, we finetune a subset of weights of SD and BLIP following heuristics~\cite{kumari2022multi} and hyperparameter search. For BLIP ViT-B and ViT-L, we finetune the query projection weights in the cross-attention layer of the text decoder and freeze the other components. For BLIP-2, the query tokens are adapted. For SD, we finetune the query projection weights in the cross-attention layer. Unless specified, we use a learning rate of 1e-4 and batch size of 8 and finetune the framework for 5 epochs. The experiments are conducted on an A10 GPU with 24GB of memory. The finetuning is efficient, requiring only a forward pass for both BLIP and SD per input. Thus, the additional computations introduced by the reconstruction process are relatively minor, amounting to approximately 1.4 times the original cost.

\begin{figure*}[htp]
    \centering
    \includegraphics[width=\textwidth]{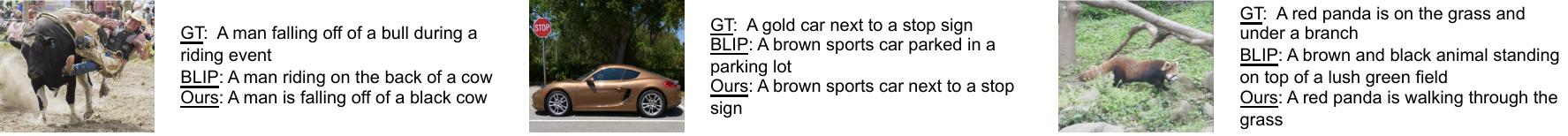}
    \caption{Qualitative evaluation of image captioning. From top to down, we show a ground-truth (GT) caption, a generated caption from BLIP ViT-B, and a generated caption from our finetuned model of ViT-B.}
    \label{fig:qualitative_caption}
\end{figure*}

\begin{figure}[htp]
    \centering
    \includegraphics[width=\linewidth]{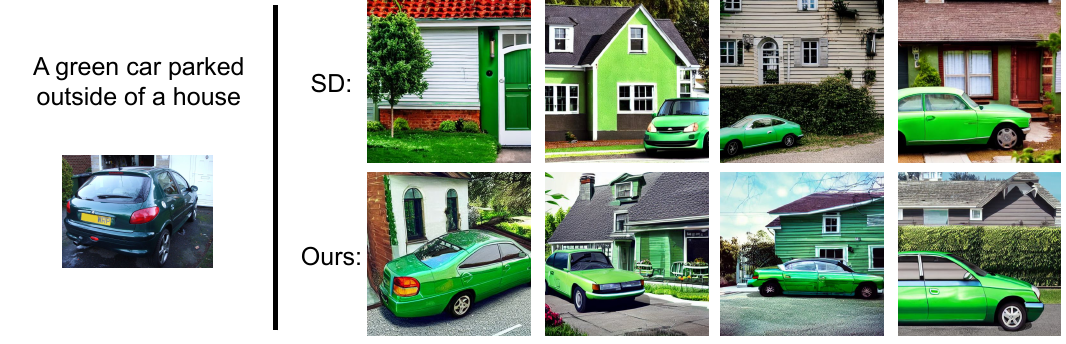}
    \caption{Qualitative evaluation of image generation. The left side shows a textual input and a reference image. The upper images on the right side are generated by the SD baseline. The lower images are generated from our model.}
    \label{fig:qualitative_image}
\end{figure}

\section{Evaluation Results}

Here we present a comprehensive evaluation of our framework by comparing it to baselines and showing qualitative results. In addition, we discuss and analyze our training objective.

\vspace{0.1cm} 
\noindent \textbf{Improvement in Image Captioning}
Table \ref{tab:image} shows the performance of different models on the image captioning task. We report a broader set of metrics on the entire dataset for a more detailed evaluation. As shown in the table, our finetuned model demonstrated improved performance across most metrics. Our approach is effective across different sizes (ViT-B vs ViT-L) and architectures (BLIP vs BLIP-2) of image captioning models. Note that our model outperforms ViT-B with a $2\%$ improvement in CIDEr, which is a larger gain compared to BLIP's $0.7\%$ improvement obtained through bootstrapping on a dataset of 129M images.

\vspace{0.1cm} 
\noindent \textbf{Improvement in Image Generation}
Similarly, we compare the performance of text-to-image generation models in Table \ref{tab:text}. 
We observe that SD finetuned with our reconstruction loss can produce more semantically aligned images on the NoCaps dataset that contains diverse and real-world scenarios, despite slightly reduced fidelity. Finetuning SD solely with MSE loss on human-labeled image-text pairs yields improvements in the COCO test set, which has the same data distribution as the training data. However, this approach leads to worse performance in image fidelity and degraded alignment on the NoCaps dataset. In general, utilizing reconstruction loss leads to improvements over human supervision. Our proposed framework has demonstrated significant potential to improve image generation.

\begin{table}[htp]
    \centering
    \scriptsize
    \resizebox{\linewidth}{!}{
    \begin{tabular}{l|cc|cc}\toprule
    & \multicolumn{2}{c}{NoCaps} & \multicolumn{2}{|c}{COCO} \\
    Model &  CLIP $\downarrow$ &FID $\downarrow$ & CLIP $\downarrow$ &FID $\downarrow$ \\\hline
    SD~\cite{rombach2022high} & 0.4039 & \textbf{21.19} & 0.4248 & 27.08\\
    SD MSE & 0.4077 & 24.32 & \textbf{0.4011} & 25.66\\ \hline
    Ours ViT-B, SD & \textbf{0.3978}& 21.96 & 0.4068 & 24.68\\
    Ours ViT-L, SD & 0.4031 & 22.92 & 0.4071 & \textbf{24.21} \\ 
    \toprule
    \end{tabular}
    }
    \caption{Evaluation results of image generation. We compare baseline methods, SD and MSE loss, with our image generation models that are finetuned with two image captioning models.}
    \label{tab:text}
\end{table}

\vspace{0.1cm} 
\noindent \textbf{Qualitative Results}
Our method generates a more faithful description of a scene in terms of relationships, context, and fine-grained concepts. For example, BLIP correctly recognizes the objects in the left-most image in Figure \ref{fig:qualitative_caption}, but it fails to predict the relationship between them. In contrast, our model generates a correct description, likely driven by the reconstruction loss that the visual perception of a man riding a cow is different from that of a man falling off a cow.
In addition, Figure \ref{fig:qualitative_image} shows an example of generated images using our model and SD baseline method. The baseline may neglect certain objects like the car, whereas our method reflects the text prompt. More figures are in Appendix \ref{app:training_qualitative}.

\begin{table}[htp]
    \centering
    \resizebox{\linewidth}{!}{
    \begin{tabular}{l|cc|l|cc}\toprule
    \multicolumn{3}{c}{Captioning} & \multicolumn{3}{|c}{Image Generation} \\ \hline
    Model &  CIDEr & SPICE & Model & CLIP $\downarrow$ &FID $\downarrow$ \\\hline
    Ours & \textbf{111.8} & \textbf{14.9} & Ours & \textbf{0.3978} & \textbf{21.96}\\
    w/o $\mathcal{L}_{TG}$ & 102.3 & 14.0 & w/o $\mathcal{L}_{IG}$& 0.4241 & 26.56 \\
    w/o $\mathcal{L}_{IR}$ & 109.7 & 14.7 & w/o $\mathcal{L}_{TR}$& 0.4077 & 24.32 \\
    \toprule
    \end{tabular}
    }
    \caption{Analysis of the loss function. We conduct experiments with either only the regularization loss or only the supervision of image-text pairs.}
    \label{tab:ablation}
\end{table}

\vspace{0.1cm} \noindent \textbf{Analysis of Loss Function}
We investigate the effectiveness of our proposed loss function, especially examining if the reconstruction loss alone can still lead to the improvement exhibited above. The first row in Table \ref{tab:ablation} shows the results of our final models trained on the combined pipelines using Eq. \ref{eq:loss}, whereas for the second and third rows, we separately train the two pipelines and ablate the loss terms. The second row on the left side of the table refers to the BLIP model trained solely on $\mathcal{L}_{IR}$ using the framework in Figure \ref{fig:loss} (a). As expected, the model struggles to provide competitive results. This shows the benefit of human supervision, which prevents the model from overfitting the reconstruction objective. For completeness, we also show the performance of BLIP with only the image captioning objective in the third row of Table \ref{tab:ablation}. As our training framework involves different loss terms and finetuning strategies, a more detailed analysis of those aspects can be found in Appendix~\ref{app:ablation}.

\section{Discussions and Limitations}
Besides our findings and improvements, this work also opens up avenues for potential future directions. 1) The novel design of backpropagating through SD allows image generation to be used as a downstream task, wherein the knowledge in diffusion models is effectively transferred. 2) The findings in our work inspire the development of label-free evaluation metrics for image captioning.

The primary limitation of this work is the restricted scope of improvement. In our work, the degree of improvement in BLIP and SD depends on their initial capacities. As a result, challenges may remain for complex images or intricate descriptions which are not well covered in the data distribution for pretraining. Besides, we treat the transformations from image to text and text to image as a black box, lacking a deeper understanding of the alignment between layers within models in the generation processes. We leave further exploration with explanation methods in future work~\cite{selvaraju2017grad,bach2015pixel,gu2019understanding}. Another under-explored perspective is the robustness of our finetuning paradigm. Concretely, it is not clear how the fine-tuned models perform under out-of-distribution images and texts~\cite{gu2023towards,chen2023benchmarking}. Additionally, our work also inherits the known limitations of large-scale generative models~\cite{alayrac2022flamingo,zhang2022opt}, bringing concerns about possible biases or harmful content generation.

\section{Conclusion}

In this work, we investigated mutual understanding between multimodal image-to-text models and text-to-image models through image-text-image and text-image-text reconstruction tasks. We found that the reconstruction quality of text-to-image and image-to-text models can be utilized to evaluate the quality of the text or image generation. Specifically, the best textual description for an image is one that leads to a better reconstruction of the input image. The best image representation of text input is one that leads to a better recovery of the original text. Leveraging these findings, we proposed a novel framework for finetuning the image captioning and image generation models. We demonstrated enhanced performance of our models on both tasks. Our work advocates further exploring multimodal communication between text-to-image and image-to-text models. Finally, our work demonstrated the value of symbolic sentences to convey information: Image content can effectively be compressed into a sentence, and a sentence can be reconstructed as an image. This latter step can be considered a form of grounded cognition or embodiment.

{\small
\bibliographystyle{ieee_fullname}
\bibliography{egbib}
}

\clearpage
\appendix

\section*{Supplementary Material}

\section{Experimental Details}
\label{app:finding_exps}

\subsection{Implementation of Similarity Metrics}
This subsection lists approaches for computing mage-to-image similarities and text-to-text similarities. In the image-text-image and text-image-text tasks, the evaluation metrics and distance metrics are closely related, i.e., the evaluation metrics for one task serve as distance metrics for the other task. Hence, we present them jointly in this section.

\vspace{0.1cm}
\noindent \textbf{CLIP:} The CLIP ViT-L/14 model\footnote{adapted from the public library \url{https://huggingface.co/docs/transformers/model doc/clip}} is pre-trained on a 400M text-image pair dataset. Specifically, we employ the outputs from the visual projection layer with an embedding size of 768 for the CLIP image encoder. For the CLIP text encoder, the outputs from the textual projection layer with an embedding size of 512 are used.

\vspace{0.1cm}
\noindent \textbf{DINO:} The pretrained DINO ViT-B/8 model\footnote{adapted from \url{https://github.com/facebookresearch/dino}} are used to obtain the image embeddings. DINO has been trained on the images dataset in a self-supervised way and has shown superior performance on representation learning tasks. The input image size is set to 384 and the output embedding size to 768.

\vspace{0.1cm}
\noindent \textbf{FID, IS:} Following~\cite{rombach2022high}, the torch-fidelity library\footnote{public implementation available from \url{https://github.com/toshas/torch-fidelity}} is used to compute the fidelity scores of the generated images.

\vspace{0.1cm}
\noindent \textbf{SBERT:} We use Sentence-BERT\footnote{public implementation from \url{https://huggingface.co/sentence-transformers/all-MiniLM-L6-v2}} with embedding size of 384 and take the [CLS] embeddings from the last transformer layer as the representation for the input text.

\vspace{0.1cm}
\noindent \textbf{WMD:} We use the open implementation of NLTK library\footnote{\url{https://www.nltk.org/}} to compute WMD between a source sentence and a target sentence. Specifically, we use the \textit{glove-wiki-gigaword} vector embeddings with 200 dimensions as the choice for the WMD. Since WMD is a distance metric rather than a score metric, we plot the $y$-axis in the reversed order to represent that the higher the $y$, the better the text, as shown in the second subfigure in Figure \ref{fig:finding_image}.

\subsection{Details of Generation Models and Benchmarks}
We use the NoCaps~\cite{agrawal2019nocaps} validation set of 4500 images and a subset of 2000 images from the COCO Karpathy test split~\cite{karpathy2015deep} to support the evaluation. For BLIP\footnote{\url{https://github.com/salesforce/BLIP}}, we use the ViT-L model finetuned for the image captioning task. For SD\footnote{\url{https://github.com/huggingface/diffusers}}, we use the weights sd-v1-4.ckpt and DDPM scheduler with sampling steps of 50 and a guidance scale of 7.5. The output image size is 512. The number of minimal words is set to 5 and the maximum number of words is set to 30. We set $p$=0.9 in the Top-$p$ strategy.

\section{Impact of Parameter of Top-$p$ Sampling}
\label{app:top_p}

\begin{table}[htp]
    \centering
    \small
    \begin{tabular}{c|ccccc}\toprule
        $p$ & 0.1 & 0.2 & 0.3 & 0.4 & 0.5  \\ \hline
        CIDEr & 111.8 & 111.2 & 110.0 & 107.2 & 103.4  \\\hline\hline
        $p$ & 0.6 & 0.7 & 0.8 & 0.9 & 0.95\\ \hline
        CIDEr & 96.2 & 89.7 & 83.1 & 78.7 & 69.2\\ \toprule
    \end{tabular}
    \caption{Impact of $p$ on the Top-$p$ sampling performance of BLIP ViT-L.}
    \label{tab:topp}
\end{table}

\begin{table*}
    \centering
    \resizebox{\textwidth}{!}{
    \scriptsize
    \begin{tabular}{l|cccccccc|cccc}\toprule
    & \multicolumn{8}{c}{Nocaps}& \multicolumn{4}{|c}{COCO} \\
    \cline{2-13}
    & \multicolumn{2}{c}{In-domain}& \multicolumn{2}{c}{Near-domain}& \multicolumn{2}{c}{Out-domain} & \multicolumn{2}{c}{Overall}& \multicolumn{4}{|c}{Karpathy Test} \\
    Method& CIDEr & SPICE  & CIDEr  & SPICE  & CIDEr  & SPICE  & CIDEr  & SPICE &B@3 & B@4 & CIDEr & SPICE \\\hline
    Top-$k$ Sampling & 79.5 & 13.2 & 78.4 & 12.4 & 83.4 & 11.7 & 79.6 & 12.4& 39.7 & 28.1 & 108.4 & 20.9\\
    Ours & \textbf{84.3} & \textbf{13.5} & \textbf{81.6} & \textbf{12.8} & \textbf{91.6} & \textbf{12.7} & \textbf{84.0} & \textbf{12.9} & \textbf{40.1} & \textbf{28.7} & \textbf{111.6} & \textbf{21.5}\\ \hline
    Gain (\%) & +6.0 & +2.3 & +4.1 & +3.2 & +9.8 & +8.5 & +5.5 & +4.0& +1.1 & +2.3 & +2.9 & +2.9\\
    \hline
    \hline
    Tempered Sampling& 83.9 & 13.2 & 82.7 & 12.7 & 89.8 & 12.0 & 84.3 & 12.6& 33.0 & 22.2 & 92.4 & 19.5\\
    Ours & \textbf{87.8} & \textbf{13.4} & \textbf{87.0} & \textbf{13.1} & \textbf{98.1} & \textbf{12.8} & \textbf{89.3} & \textbf{13.1}& \textbf{33.6} & \textbf{22.8} & \textbf{95.4} & \textbf{20.4}\\ \hline
    Gain (\%) & +4.6 & +1.5 & +5.2 & +3.1 & +9.2 & +6.7 & +5.9 & +4.0& +1.8 & +2.6 & +3.2 & +4.7 \\\toprule
    \end{tabular}
    }
    \caption{Comparison of different sampling methods and our proposed method on Nocaps and COCO datasets. Our method outperforms every sampling method on all metrics. The relative gain of our method compared to each sampling method is given in the last row in each block. B@$k$: BLEU@$k$.}
    \label{app:finding1}
\end{table*}

Top-$p$ sampling, which is also referred to as nucleus sampling, is a text generation method that samples words from a set of candidates whose cumulative probability exceeds a specified threshold $p$. By varying $p$, we achieve a trade-off between the diversity and accuracy of the generated text. Broadly speaking, larger values of $p$ result in more diverse captions, whereas smaller values of $p$ lead to less variable yet more accurate captions for an input image. 

Note that in Table 2 of the BLIP paper~\cite{li2022blip}, the Top-$p$ sampling method is used to generate a diverse set of captions which are then utilized for bootstrapping the BLIP model. However, the evaluation result in that row is obtained by the beam search method. To examine the effect of $p$, we report the performance of BLIP ViT-L on the NoCaps dataset for image captioning, under different choices of $p$, in Table \ref{tab:topp}.

\section{Influence of Sampling Methods for Image Captioning}
\label{app:sampling_methods}
Several sampling methods exist for generating text in the image captioning model. Apart from the Top-$p$ sampling approach presented in the main text, we conduct a quantitative evaluation of two different sampling strategies, including Top-$k$~\cite{fan2018hierarchical} with $k$=10, and Tempered sampling~\cite{caccia2018language, nadeem2020systematic} with $T$=0.7. Each sampling method serves as a baseline. Table \ref{app:finding1} verifies that our conclusion is solid across different sampling methods.

\section{Impact of Number of Candidates $N$}
We investigate the effect of the number of candidates $N$ on our findings. We conduct experiments using the BLIP ViT-L model on the NoCaps dataset for image captioning, utilizing the Top-$p$ sampling method. For image generation, we use the SD model and the DDPM scheduler. We sample $N$ candidates for each input image or text in each experiment. We compare our approach to the baseline method, where a candidate is selected randomly.

\label{app:finding_num}
\begin{figure}[htp]
    \centering
    \includegraphics[width=0.85\linewidth]{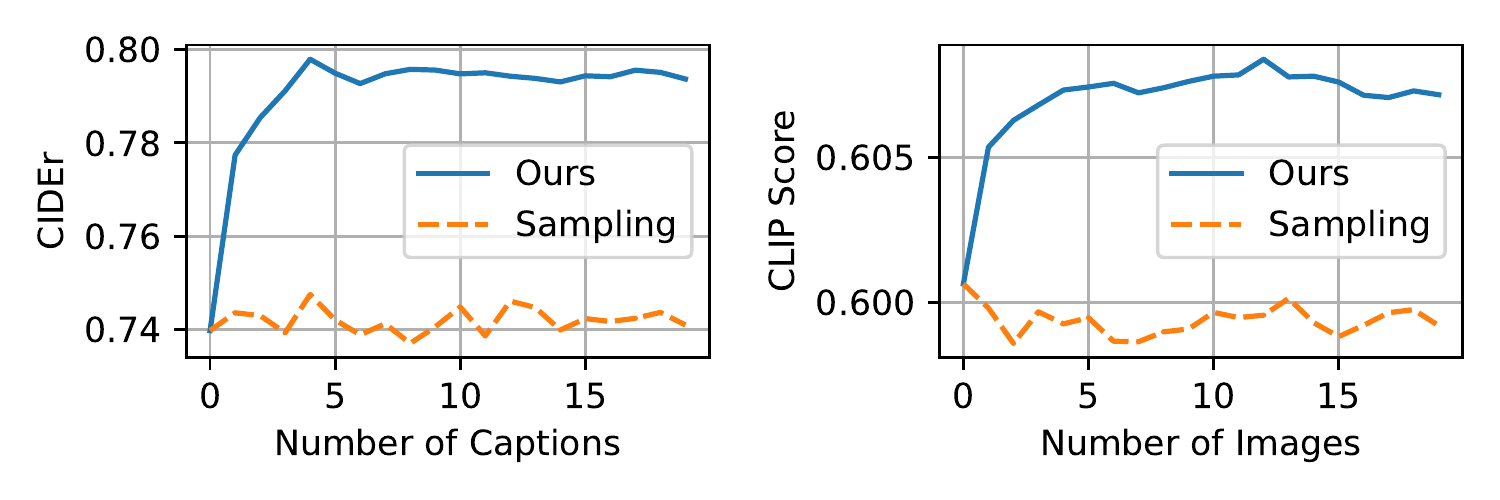}
    \caption{Evaluation of the choice of the number of sample candidates.}
    \label{fig:finding_hyperparam}
\end{figure}

Figure \ref{fig:finding_hyperparam} depicts the impact of the number of candidates on image-text-image (left) and text-image-text (right) tasks. The figure (left) shows the image captioning score ($y$-axis) for our approach and the baseline method, with the $x$-axis representing the number of candidate captions for each input image. As shown in the figure, the performance of the baseline method remains consistent as it randomly samples captions with varying qualities. Conversely, our approach improves significantly after the number of captions has reached around five. When the number of candidates is limited, there may not be enough high-quality captions to choose from, resulting in lower performance. Nevertheless, our approach remains effective even in that stage. After acquiring a reasonable number of candidates, our method consistently outperforms the baseline method by a significant margin. A similar conclusion can be inferred for the text-image-text task, as demonstrated on the right side of the figure.

\section{Qualitative Results for Image and Text Generation}
\label{app:finding_qualitative}
\begin{figure*}[]
    \centering
    \scriptsize
    \setlength{\tabcolsep}{0pt}
    \begin{tabular}{l}
        \toprule
        \hspace{0.2in}Input \hspace{0.75in} Sample \#1 \hspace{0.75in}  Sample \#2 \hspace{0.75in}  Sample \#3 \hspace{0.75in}  Sample \#4 \hspace{0.75in}  Sample \#5\\\hline \\[-1.5ex]
        \includegraphics[width=\textwidth]{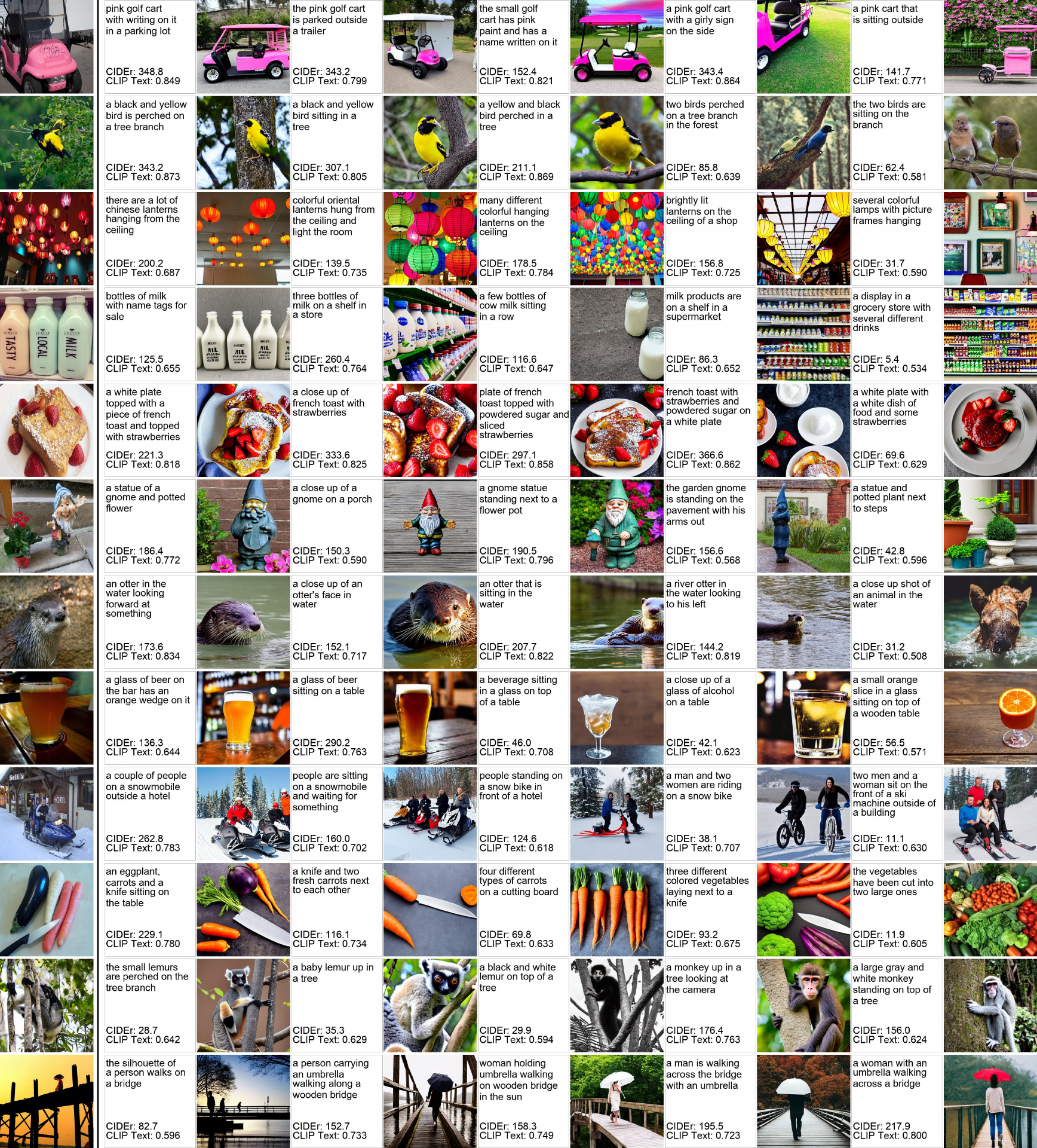}\\
    \end{tabular}
    \caption{Examples for the image-text-image task using Top-$p$ sampling. The first column displays the input images, followed by the generated caption and its corresponding generated image. We rank the generated text-image pairs based on the similarity of the images and show the score of the caption below each text. In the first row, \textit{golf car} in the first sample is a more accurate description than \textit{cart} in the fifth sample so that the first generated image is closer to the input image. Additionally, we show a few failed examples in the last two rows.}
    \label{fig:image_positive}
\end{figure*}

\begin{figure*}[]
    \centering
    \scriptsize
    \setlength{\tabcolsep}{0pt}
    \begin{tabular}{l}
        \toprule
        \hspace{0.2in}Input \hspace{0.75in} Sample \#1 \hspace{0.75in}  Sample \#2 \hspace{0.75in}  Sample \#3 \hspace{0.75in}  Sample \#4 \hspace{0.75in}  Sample \#5\\\hline \\[-1.5ex]
        \includegraphics[width=\textwidth]{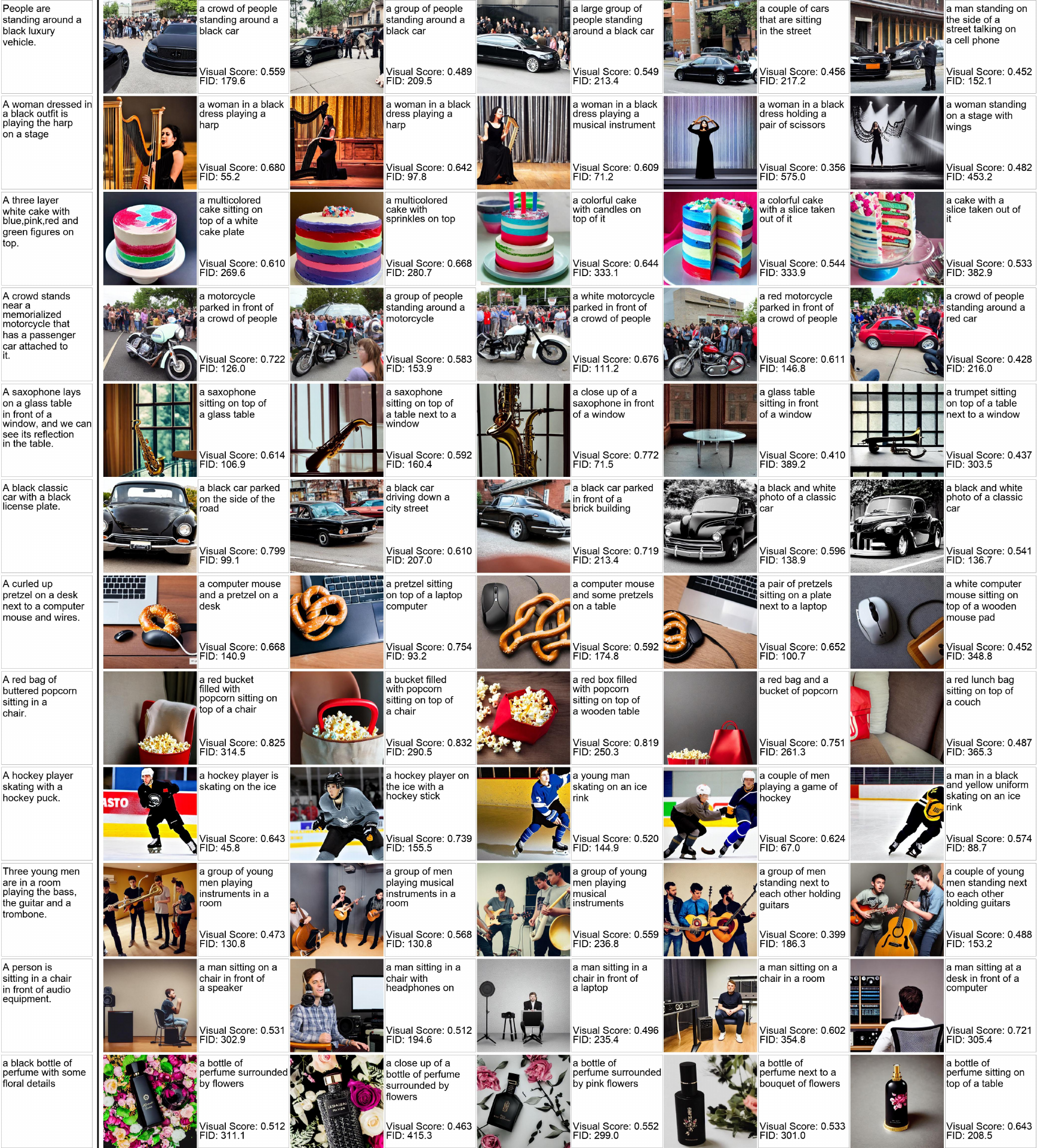}\\
    \end{tabular}
    \caption{Examples for the text-image-text task. The first column shows the input text, followed by generated image and its corresponding generated text. We rank the generated image-text pairs by the similarity of the text and show the score of the image in the box. As shown in the first line, the image in the first sample represents the input text better than the image in the fifth sample, and the similarity of the reconstructed text reflected this comparison. Further, we show some failed examples in the last two rows.}
    \label{fig:text_positive}
\end{figure*}

To reinforce the findings in Section \ref{sec:finding}, we provide additional qualitative examples on the NoCaps dataset. The annotations and explanations for each figure are included in their respective captions. Figure \ref{fig:image_positive} shows both positive examples and negative examples for the image-text-image task, whereas Figure \ref{fig:text_positive} presents visualizations for the text-image-text task. All results are obtained from BLIP ViT-L and SD models.

\section{Analysis of Different Image-to-Text and Text-to-Image Generative Models}
\label{app:diff_models}
We conduct further experiments with different image captioning and text-to-image models. Table \ref{tab:finding1_models} shows the result of BLIP with a VAE-based image generative model LAFITE for the image-text-image task, as well as SD with BLIP-2 for the text-image-text task. In general, our finding, that better reconstruction leads to better generation performance, still holds. We find that SD performs better than LAFITE, very likely due to its larger training data. When coupled with BLIP-2, SD improved the performance upon the baseline, but it performed worse than that in the case of BLIP.

\begin{table}[htp]
    \centering
    \small
    \resizebox{0.95\linewidth}{!}{
    \begin{tabular}{ll|cccc}\toprule
    \multicolumn{6}{c}{Text Generation}\\ \hline
    I2T & T2I &  I-C & N-C & O-C & E-C  \\\hline
    \multicolumn{2}{l|}{Baseline} & 75.1 & 72.4 & 78.7 & 74.1\\
    BLIP~\cite{li2022blip} & SD~\cite{rombach2022high} & \textbf{77.3} & \textbf{78.3} & \textbf{88.8} & \textbf{80.3}   \\ 
    BLIP~\cite{li2022blip} & LAFITE~\cite{zhou2021lafite} & 75.4 & 76.2  & 82.9 &  77.4  \\ \hline
    \multicolumn{6}{c}{Image Generation}\\ \hline
    T2I & I2T & CLIP$\downarrow$ & FID$\downarrow$  & \multicolumn{2}{c}{IS$\uparrow$ }  \\\hline
    \multicolumn{2}{l|}{Baseline} & 40.54 & 32.37 & \multicolumn{2}{c}{41.19 $\pm$ 2.91}\\
    SD~\cite{rombach2022high} & BLIP~\cite{li2022blip} & \textbf{33.47} & \textbf{29.59} & \multicolumn{2}{c}{\textbf{45.64} $\pm$ 2.40} \\ 
    SD~\cite{rombach2022high} & BLIP-2~\cite{li2023blip} & 39.41 & 31.34 & \multicolumn{2}{c}{42.34 $\pm$ 2.23} \\ 
    \toprule
    \end{tabular}
    }
    \caption{Comparison of different combinations of generation models. We evaluate two image captioning models and two image generation models on the NoCaps dataset. I2T: Image-to-text model. T2I: Text-to-image model. I-C/N-C/O-C/E-C: In-/Near-/Out-/Entire-domain CIDEr.}
    \label{tab:finding1_models}
\end{table}

\section{Implementations of Tokenizer Transformation}
\label{app:training_tokenizer}
We elaborate on the gradient backpropagation process between the output of BLIP and the input of SD, shown on the left side of Figure \ref{fig:loss}. Our framework includes three types of text tokenizers: BLIP utilizes the word-piece tokenizing method; BLIP-2 uses the byte-pair-encoding tokenizing method; SD employs the CLIP tokenizer, which uses the word-level tokenizing strategy. One of the challenges we faced is to align the output token distributions from BLIP with those from SD, allowing SD to interpret the output of BLIP. To address this issue, we use these tokenization strategies to tokenize each sentence in the COCO training set and learn a one-to-one hard-coded mapping from a source token to a target token. Despite using different strategies, we found that these tokenizing methods have a high ratio of overlapping between tokens. Specifically, when applied to the COCO training set, more than 60\% captions can be tokenized into the same set of tokens for BLIP and SD. For unmatched tokens, we map them to the most similar ones or to the [UNK] token. We visually examined this method by generating images conditioned on token distributions and found it to be practical. Additionally, we remove the prefix tokens of BLIP, \textit{a photo of}, and add [BOS] and [EOS] tokens for SD.

\section{Discussion on the Loss Function}
\label{app:pseudo_code}

\vspace{0.1cm}\noindent\textbf{Parameter Update} While optimizing $\mathcal{L}_{IR}$ for image captioning, both SD and BLIP are trained. $\mathcal{L}_{TG}$ only updates the parameters of BLIP. Likewise, both SD and BLIP are trained when optimizing $\mathcal{L}_{TR}$, and only SD is updated by $\mathcal{L}_{IG}$. The reasons for training SD in $\mathcal{L}_{IR}$ are twofold. First, SD needs to adapt to new distributions coming from BLIP. As in the standard training, the input of BLIP is discrete tokens. Whereas in our approach, the input is token distributions. Second, SD could be improved because of additional training data sampled from BLIP. In addition, since they are refreshed at each iteration, one model is able to provide better samples to train the other model throughout the training.

\vspace{0.1cm}\noindent\textbf{Connection to CycleGAN} CycleGAN is a generative model that learns bidirectional mappings from domain $X$ to domain $Y$, where $X$ and $Y$ are images. Its cyclic loss ensures that the mapping between the input and output domains is consistent, i.e., if we take an image from domain $X$, pass it through the generator network to obtain an image in domain $Y$, and then pass that image through the generator network again to obtain an image in domain $X$, we should obtain an image that is similar to the original image in domain $X$. Likewise, we aim to enforce consistent mapping between the input and output domains, but we deal with two distinct domains, i.e., image and language, which may require much more complex mapping. In addition, we also optimize a single objective, similar to CycleGAN which is trained on the weighted cyclic loss and adversarial loss. CycleGAN employs a hyperparameter $\lambda$ to control the relative importance of the cyclic loss to the adversarial loss, we found that a similar weighting did not yield substantial differences in performance in our work. Therefore we do not adjust this hyperparameter in our approach.

\vspace{0.1cm}\noindent\textbf{Pseudo-Code} The pseudo-code of our train framework is presented in Algorithm \ref{alg:training}.

\begin{algorithm}[htp!]
  \caption{Training Framework} \label{alg:training}
  \small
  \hspace*{\algorithmicindent} \textbf{Model} UNet $\boldsymbol{\epsilon}_\psi$, SD's Text Encoder $\pi$, BLIP $b_\theta$\\
  \hspace*{\algorithmicindent} \textbf{Input} an image-text pair $(\mathbf{x}_0, \mathbf{y})$
  \begin{algorithmic}[1]
    \Repeat
      \Statex \textcolor{blue}{\# Image-Text-Image (BLIP $\rightarrow$ SD)}
      \State $\mathbf{x}_0, \mathbf{y} \sim q(\mathbf{x}_0, \mathbf{y})$ \Comment{Sample an image-text pair from the dataset}
      \State $\hat{\mathbf{y}} = b_\theta(\mathbf{x}_0, \tilde{\mathbf{y}})$ \Comment{$\hat{\mathbf{y}} \in \mathbb{R}^{L \times V}$: Output token distribution. $\tilde{\mathbf{y}}$: (causal) masked text input}
      \State $\mathcal{L}_{1} = \mathrm{CE}(\mathbf{y}, \hat{\mathbf{y}})$ \Comment{CE: cross entropy loss}
      \State $\mathbf{c} = \pi(\hat{\mathbf{y}})$ \Comment{Text encoder encodes BLIP's output into embeddings}
      \State $t \sim \mathrm{Uniform}(\{1, \cdots, T\})$ \Comment{Sample a timestep for the diffusion process}
      \State $\boldsymbol{\epsilon}\sim\mathcal{N}(\textbf{0},\textbf{I})$ \Comment{Sample noise for timestep $t$}
      \State $\mathbf{x}_t = \sqrt{\alpha_t} \mathbf{x}_0 + \sqrt{1-\alpha_t}\boldsymbol{\epsilon}$ \Comment{Add noise to image}
      \State $\hat{\boldsymbol{\epsilon}} = \boldsymbol{\epsilon}_\psi(\mathbf{x}_t, t, \mathbf{c})$ \Comment{UNet predicts noise $\hat{\boldsymbol{\epsilon}}$ from the noisy image $\mathbf{x}_t$}
      \State $\mathcal{L}_2 = \left\| \boldsymbol{\epsilon} -  \hat{\boldsymbol{\epsilon}} \right\|^2$
      
      \Statex \textcolor{blue}{\# Text-Image-Text (SD $\rightarrow$ BLIP)}
      \State $\mathbf{x}_0, \mathbf{y} \sim q(\mathbf{x}_0, \mathbf{y})$ \Comment{Sample an image-text pair from the dataset}
      \State $\mathbf{c} = \pi(\mathbf{y})$ \Comment{Text encoder encodes input text into embeddings}
      \State $t \sim \mathrm{Uniform}(\{1, \cdots, T\})$ \Comment{Sample a timestep for the diffusion process}
      \State $\boldsymbol{\epsilon}\sim\mathcal{N}(\textbf{0},\textbf{I})$ \Comment{Sample noise for timestep $t$}
      \State $\mathbf{x}_t = \sqrt{\alpha_t} \mathbf{x}_0 + \sqrt{1-\alpha_t}\boldsymbol{\epsilon}$ \Comment{Add noise to image}
      \State $\hat{\boldsymbol{\epsilon}} = \boldsymbol{\epsilon}_\psi(\mathbf{x}_t, t, \mathbf{c})$ \Comment{UNet predicts noise $\hat{\boldsymbol{\epsilon}}$ from the noisy image $\mathbf{x}_t$}
      \State $\mathcal{L}_3 = \left\| \boldsymbol{\epsilon} -  \hat{\boldsymbol{\epsilon}} \right\|^2$
      \State $\mathbf{\hat{x}}_0 = \frac{1}{\sqrt{\alpha_t}}(\mathbf{x}_t - \sqrt{1-\alpha_t}\boldsymbol{\hat{\epsilon}})$ \Comment{1-step approximation of $\mathbf{x}_0$}
      \State $\mathcal{L}_4 = \mathrm{CE}(\mathbf{y}, b_\theta(\mathbf{\hat{x}}_0, \tilde{\mathbf{y}}))$ \Comment{$\tilde{\mathbf{y}}$: (causal) masked text input}
      \State Take gradient descent step on
      \Statex \qquad BLIP:  $ \nabla_{\theta} (\mathcal{L}_1 +  \mathcal{L}_2 +  \mathcal{L}_3 +  \mathcal{L}_4) = \nabla_{\theta} (\mathcal{L}_1 +  \mathcal{L}_2 + \mathcal{L}_4)$
      \Statex \qquad SD: $ \nabla_{\psi} (\mathcal{L}_1 +  \mathcal{L}_2 +  \mathcal{L}_3 +  \mathcal{L}_4)= \nabla_{\psi} (\mathcal{L}_2 +  \mathcal{L}_3 + \mathcal{L}_4)$
      \Until{converged}\\
      \Return{$\psi, \theta$}
  \end{algorithmic}
\end{algorithm}

\section{Ablation Study}
\label{app:ablation}

\begin{table*}[htp]
    \centering
    \resizebox{0.9\linewidth}{!}{
    \begin{tabular}{cc|cc|cc|cc|c|c|c}\hline
        \multicolumn{4}{c|}{Weights Update} & \multicolumn{4}{c|}{Loss Function}&& \\ 
        $\mathrm{BLIP}^{p_1}$ & $\mathrm{SD}^{p_1}$ & $\mathrm{SD}^{p_2}$ & $\mathrm{BLIP}^{p_2}$ & $L_{TG}$ & $L_{IR}$ & $L_{IG}$ & $L_{TR}$ &  Effect on BLIP & Experiment & CIDEr \\ \hline
        $\checkmark$ & $\checkmark$ & $\checkmark$ & $\checkmark$ &$\checkmark$ & $\checkmark$ & $\checkmark$ & $\checkmark$  &  $p_1$: ground truth + reconstruction $p_2$: augmentation & Ours, Tab. 4 & 111.8 \\ \hline

        $\checkmark$ & $\times$ & $\times$ & $\times$ & $\checkmark$ & $\times$ & - & -  & $p_1$: ground truth & Tab. 4 & 109.7 \\ \hline
        
        $\checkmark$ & $\times$ & $\checkmark$ & $\times$ &$\checkmark$ & $\checkmark$ & $\checkmark$ & $\checkmark$  &  $p_1$: ground truth + reconstruction & Ablation & 110.9 \\ \hline
        $\times$ & $\checkmark$ & $\times$ & $\checkmark$ &  - & $\checkmark$ & - & $\checkmark$  & $p_2$: augmentation & Ablation & 110.2 \\ \hline

        $\checkmark$ & $\times$ & $\times$ & $\times$ &$\times$ & $\checkmark$ & - & -  &  $p_1$: reconstruction & Tab. 6 & 102.3 \\ \hline

    \end{tabular}
    }
    \caption{Analysis of different training paradigms of loss terms and model frozen strategies.}
    \label{tab:app_loss_ablation}
\end{table*}

Table \ref{tab:app_loss_ablation} summarizes the different losses and weight-freezing strategies, highlighting that improvement comes from the proposed reconstruction loss. For simplicity, we only list the effect of different settings on BLIP. Our default setting jointly optimizes both pipelines with both models trainable $p_1: \mathbf{I} \xrightarrow{\mathrm{BLIP}} \mathbf{T}(L_{TG}) \xrightarrow{\mathrm{SD}} \mathbf{I}{(L_{IR})}$, 
and $p_2: \mathbf{T} \xrightarrow{\mathrm{SD}} \mathbf{I}{(L_{IG})} \xrightarrow{\mathrm{BLIP}} \mathbf{T}{(L_{TR})}$. The third column in the table shows the training signals that BLIP can receive from the two pipelines under the specific setting. "-" means that loss has no effect as the model is frozen.

\section{Qualitative Results for Training Framework}
\label{app:training_qualitative}
\begin{figure*}
    \centering
    \includegraphics[width=\textwidth]{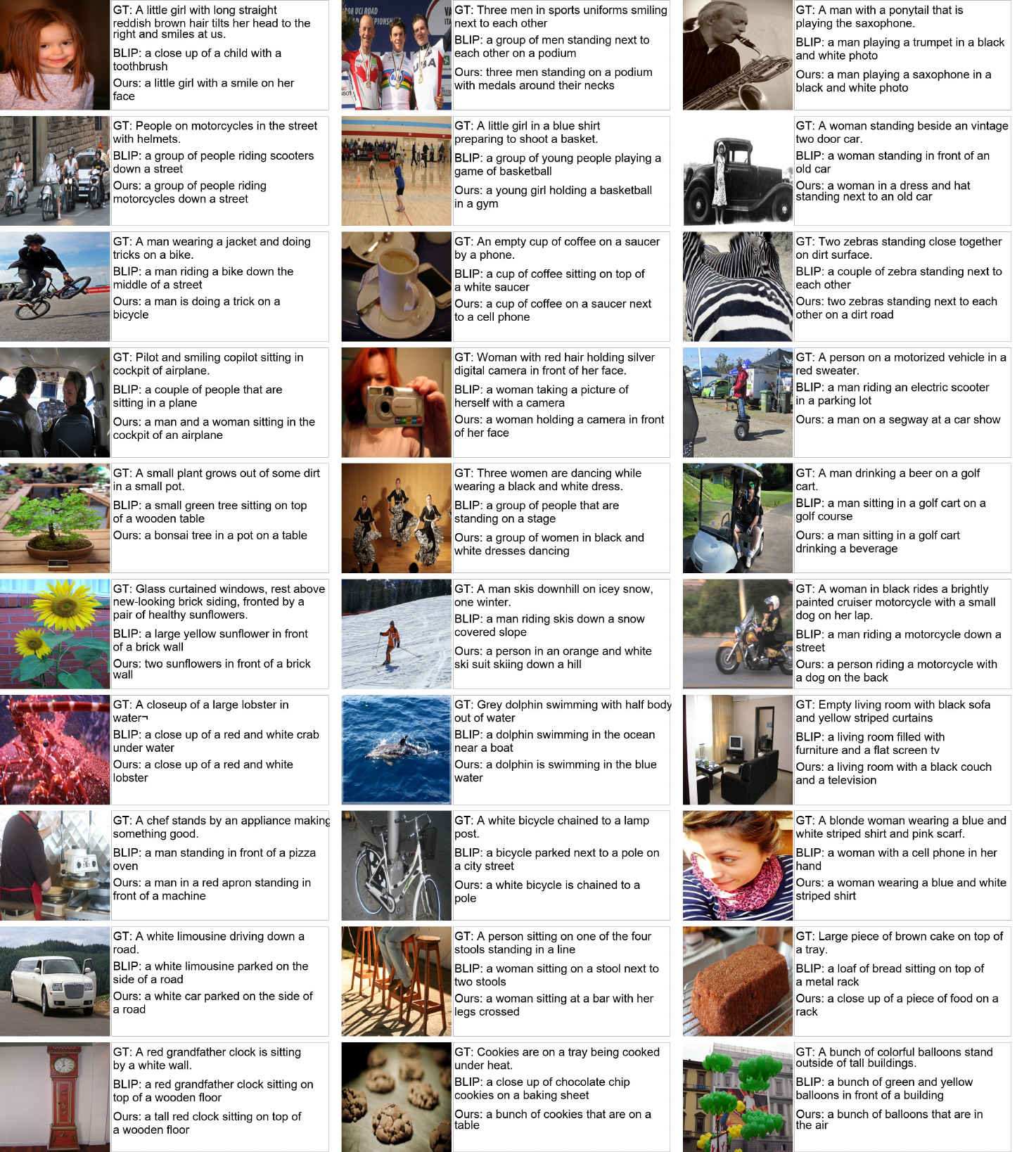}
    \caption{Qualitative results for image captioning. We compare the performance of our approach and BLIP ViT-B baseline. A ground truth (GT) caption is given for each image. On average, our method provides more accurate descriptions for input images. The last two rows show examples of our model performing worse than the baseline method.}
    \label{fig:blip_positive}
\end{figure*}

\begin{figure*}[]
    \centering
    \scriptsize
    \setlength{\tabcolsep}{0pt}
    \begin{tabular}{l}
        \toprule
        \hspace{0.2in}Input \hspace{1.8in} Ours \hspace{2.8in}  SD \\\hline \\[-1.5ex]
        \includegraphics[width=\textwidth]{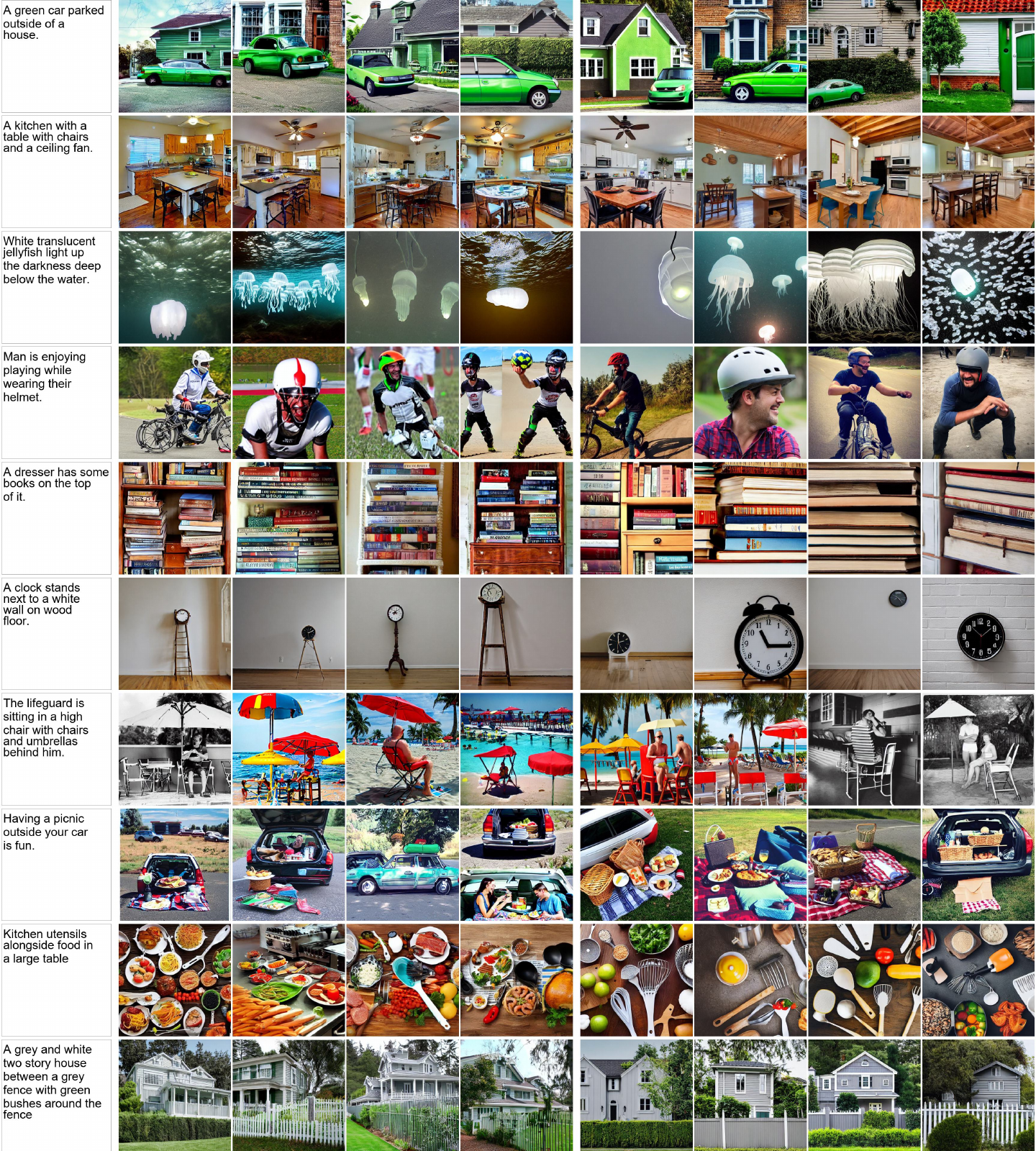}\\
    \end{tabular}
    \caption{Qualitative results for image generation. We compare the original SD model with our finetuned model. Each row displays firstly the input text, followed by four images generated by our approach, and four images generated by the SD baseline. We can see that compared with the baseline method, the image generated by our method better reflects the semantics of the input text.}
    \label{fig:sd_positive}
\end{figure*}

We provide additional qualitative examples of image captioning and image generation by our trained framework in Figure \ref{fig:blip_positive} and Figure \ref{fig:sd_positive}, respectively. Detailed annotations and explanations can be found in the corresponding figure captions.

\end{document}